\documentclass[letterpaper,10pt,conference]{ieeeconf}
\IEEEoverridecommandlockouts
\usepackage{graphicx}
\usepackage{booktabs}
\usepackage{float}
\usepackage{multicol}
\usepackage{subcaption}
\usepackage{authblk}
\usepackage{multirow}
\usepackage{amsfonts}
\usepackage{tabularx,booktabs}
\usepackage{pgfplots}
\usepackage[backend=biber,
            style=ieee]{biblatex}
\usepackage{amsmath}
\usepackage{todonotes}
\usepackage[T1]{fontenc}
\usepackage{caption}
\usepackage{tabularx}
\usepackage{booktabs,tabularx, pifont}

\makeatletter
\let\NAT@parse\undefined
\makeatother
\usepackage{hyperref}
\hypersetup{
  hypertex=true,
  colorlinks=true,
  linkcolor=blue,
  anchorcolor=blue,
  citecolor=blue
}

\title{\LARGE \bf
Tenma: Robust Cross-Embodiment Robot Manipulation with Diffusion Transformer
}

\author{
 Travis Davies$^{1}$, Yiqi Huang$^{1}$, Yunxin Liu$^{2}$, Xiang Chen$^{3}$, Huxian Liu$^{1}$, Luhui Hu$^{1}$ \\
  $^{1}$ZhiCheng AI, $^{2}$Tsinghua University, $^{3}$Peking University
}

\begin{document}
\maketitle
\thispagestyle{empty}
\pagestyle{empty}


\begin{abstract}
Scaling Transformer policies and diffusion models has advanced robotic manipulation, yet combining these techniques in \emph{lightweight}, cross-embodiment \emph{learning} settings remains challenging. We study design choices that most affect stability and performance for diffusion–transformer policies trained on heterogeneous, multimodal robot data, and introduce \textbf{Tenma}, a lightweight diffusion–transformer for bi-manual arm control. Tenma integrates multiview RGB, proprioception, and language via a cross-embodiment normalizer that maps disparate state/action spaces into a shared latent space; a \emph{Joint State–Time} encoder for temporally aligned observation learning with inference speed boosts; and a diffusion action decoder optimized training stability and learning capacity. Across benchmarks and under matched compute, Tenma achieves an average success rate of $88.95\%$ in-distribution and maintains strong performance under object and scene shifts, substantially exceeding baseline policies whose best in-distribution average is $18.12\%$. Despite using moderate data scale, Tenma delivers robust manipulation and generalization, indicating the great potential for multimodal and cross-embodiment learning strategies for further augmenting the capacity of transformer-based imitation learning policies.
\end{abstract}

\section{Introduction}
Transformer-based, imitation learning policies have recently made impressive strides in fine-grained robotic arm manipulation, with scaling behavior akin to that of large language models-where performance improves considerably with more data and model capacity \cite{pi0, rdt, octo, rt-2, lbm, openvla, llama, gpt}. With millions of robot expert demonstration trajectories now openly available, there is unprecedented potential to develop generalist policies that unify perception, language understanding, and action synthesis across diverse tasks and embodiments \cite{oxe, agibot, droid}.

\begin{figure}[!t]
  \centering
  \includegraphics[width=\linewidth]{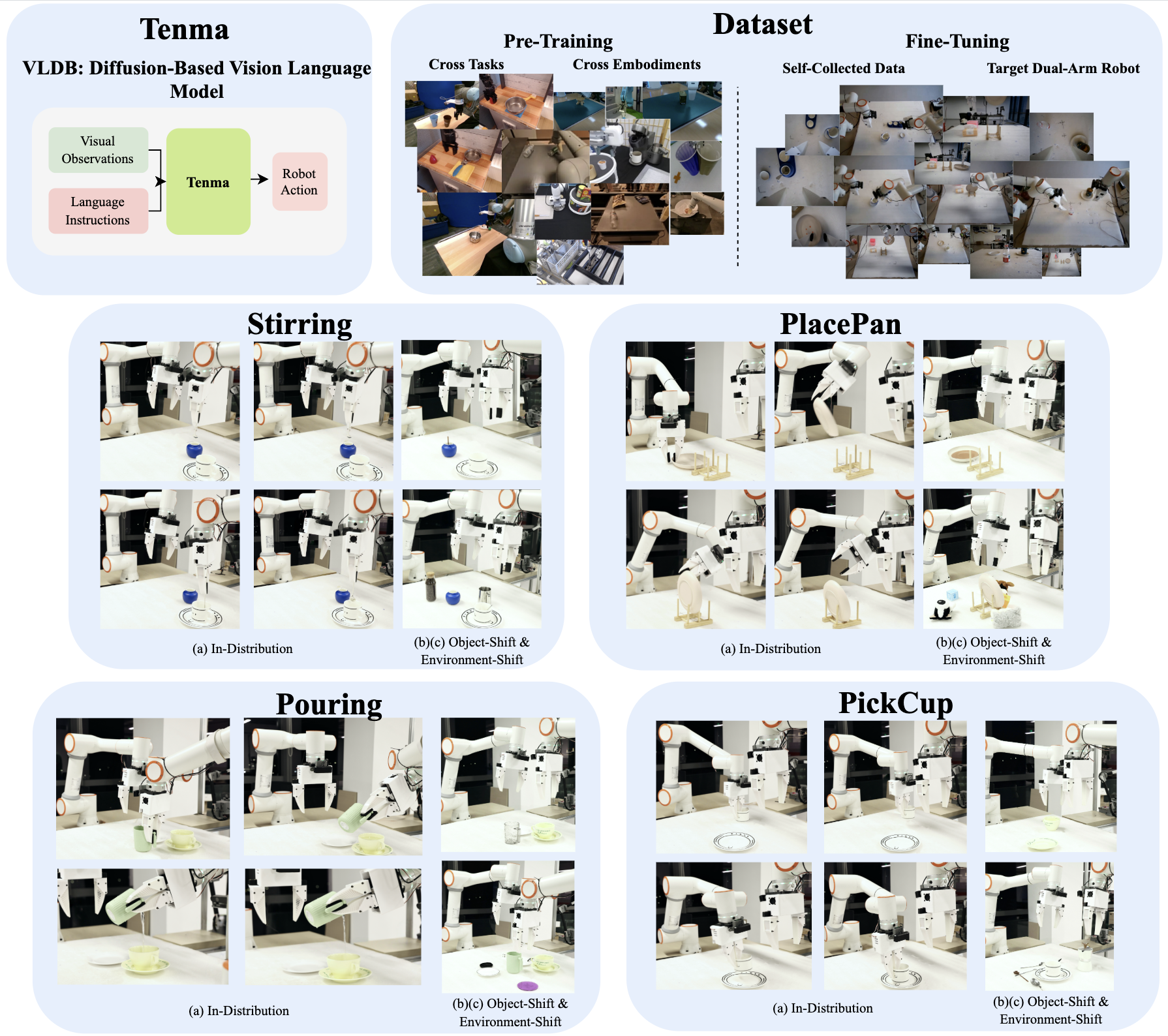}
  \caption{\textbf{Overview.} Tenma is a diffusion–transformer that learns cross-embodiment representations for robot action generation. The model is pretrained on a curated subset of Open X-Embodiment, spanning diverse robots and tasks, and fine-tuned on four dual-arm tabletop tasks: \textit{Stirring}, \textit{Pouring}, \textit{PlacePan}, and \textit{PickCup}. Evaluation considers three axes of generalization: (a) in-distribution, (b) object-shift, and (c) scene-shift. For each task, the left panels illustrate in-distribution executions, while the top-right and bottom-right panels show performance under object-shift and scene-shift conditions, respectively.
}
  \label{fig:overview}
\end{figure}

Robotic data introduces two layers of complexity absent in language‑only corpora. \textbf{First, multimodal learning}: each training sample combines five distinct modalities with mismatched structures. These include spatio‑temporal multiview RGB video, temporal robot proprioception and control signals (i.e. robot actions), and static language prompts alongside diffusion timesteps. Building high‑capacity, high‑throughput models that can \emph{stably} capture fine‑grained features from this heterogeneous data remains a major challenge in robot learning. Current approaches tackle this in different ways: some collapse the rich, high‑dimensional inputs into a single token or a token‑per‑timestep representation, sacrificing crucial details \cite{dit-policy, dp, lbm}. Others apply full self‑attention across all modalities, which is computationally expensive and risks stronger signals drowning out weaker ones \cite{octo, fp3}. In general, lightweight models tend to under-utilize pretrained ViT-based encoders, limiting their potential for rich visual representation learning, while over-relying on simple attention mechanisms for extracting observation features and using vanilla cross-attention to condition actions on these features. 

\begin{figure*}[t]
  \centering
  \includegraphics[width=0.65\textwidth]{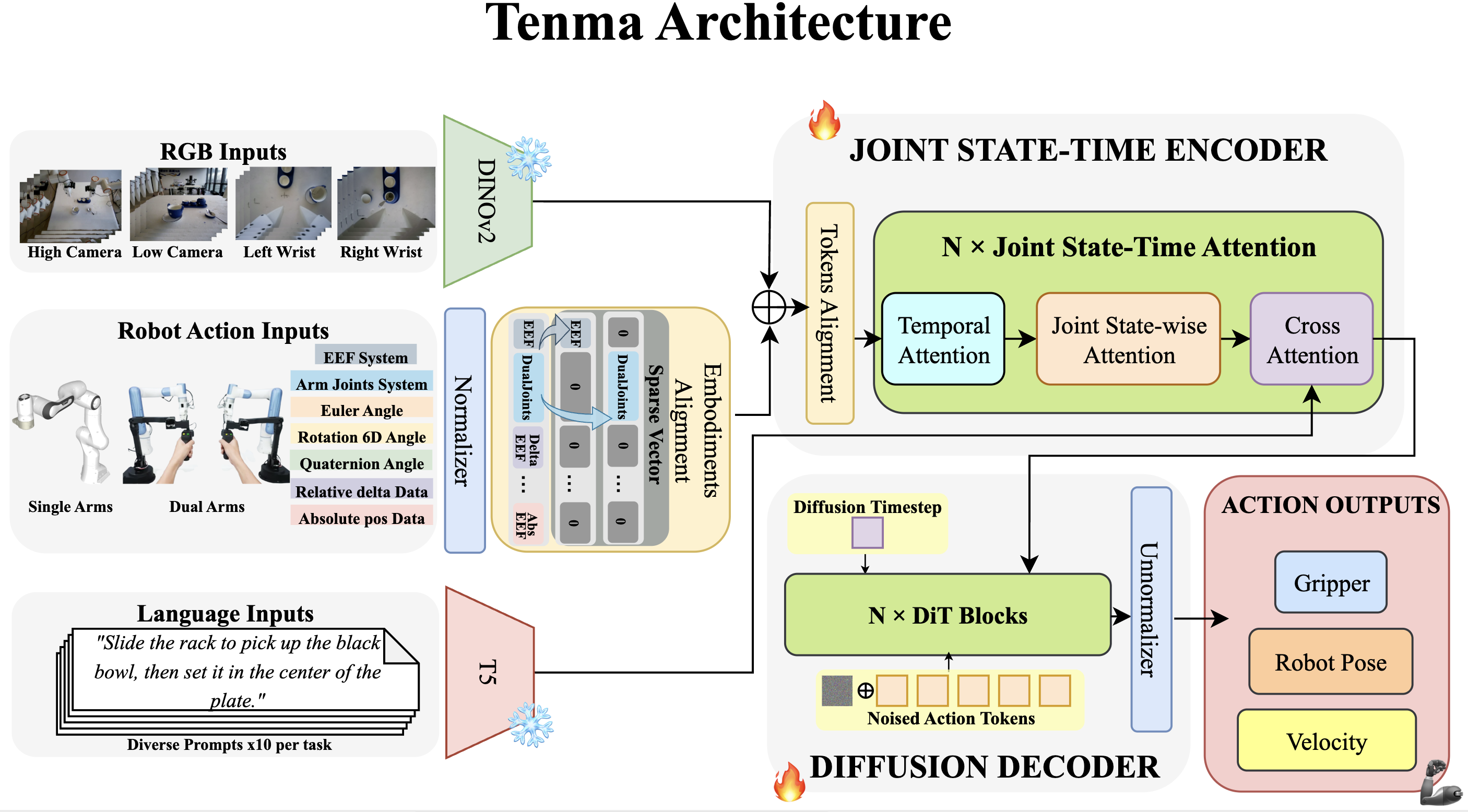}
  \caption{Overview of the Tenma architecture. The left panel depicts \textbf{cross-embodiment normalizer} and frozen multimodal \textbf{tokenizers} for vision, language, and proprioception. The top-right module represents the \textbf{Joint State–Time Encoder}, which integrates temporally aligned observation tokens with language conditioning. The bottom-right module shows the \textbf{DiT-based decoder}, which iteratively denoises action tokens, while the final block illustrates the predicted action sequence structure.}
  \label{fig:main}
\end{figure*}

\textbf{Second, cross‑embodiment heterogeneity}: Robotic platforms differ widely in their control representations, from joint angles and quaternions to end-effector poses, and span both uni-manual and bi-manual configurations. A truly generalist policy must therefore map these highly heterogeneous embodiment-specific signals into a unified latent representation and structure, enabling knowledge transfer across platforms rather than interference. Despite recent progress, this remains an unsolved challenge. Many existing models are pretrained or can only be trained on a single embodiment, such as a uni‑manual arm \cite{vjepa2, fp3, dp, dit-policy}. Others disregard critical details such as the ordering and dimensionality of RGB, proprioceptive, and action vectors \cite{octo, openvla}. An alternative is to assign each embodiment its own token projector to embed data into a shared latent space, but this design scales linearly in spatial complexity with the number of supported embodiments, creating significant overhead \cite{gr00t, pi0}.

We present \textbf{Tenma}, a lightweight diffusion-transformer architecture that learns robust features from cross-embodiment platforms to generate actions in diverse environments. Our work demonstrates that our model can complete tasks in complex settings on a given bi-manual robot arm. Our contributions are as follows:
\begin{itemize}
    \item A new lightweight transformer model which outperforms previous state-of-the art models across complex bi-manual robot arm tasks.
    \item New robot learning techniques for robust features from highly multimodal, cross-embodiment data while maintaining high inference frequency and stability. These include our Joint State-Time Attention for flexible and low-latency observation learning across embodiments, our data standardization techniques for embodiment unification, and our hybrid-conditioning diffusion decoder architecture for stable action sampling.
    \item A detailed evaluation of state-of-the-art models across a set of diverse challenges, with metrics that test abilities beyond simply success rate.
\end{itemize}

\section{Related Work}

\textbf{Scaling robot demonstration data.} Inspired by the scaling trends in natural language processing, robotics has similarly pursued large-scale demonstration learning. However, real-world robot data collection remains expensive and logistically difficult \cite{umi}. Two primary strategies have emerged: (i) open-source, real-world datasets such as DROID \cite{droid}, Open X-Embodiment \cite{oxe}, and AgiBot-World \cite{agibot}, which fuel generalist policies like Pi0 \cite{pi0} and LBM \cite{lbm}; and (ii) simulation-based pipelines leveraging environments like MuJoCo \cite{mujoco}, Isaac Gym \cite{isaacgym}, and Isaac Sim \cite{isaacsim}, as adopted by models like Gr00t-N1 \cite{gr00t}. Despite these efforts, learning from diverse datasets remains challenging due to heterogeneity in robot embodiments, sensor modalities, and camera configurations \cite{oxe}. These inconsistencies hinder the ability of models to learn a unified representation across tasks and platforms. Furthermore, established lightweight policies often lack systematic strategies for handling embodiment diversity at the data level \cite{dp, octo, dit-policy, act}. In this work, we leverage a diverse set of Open X-Embodiment robot demonstrations to learn cross-embodiment features, as well as introduce a data standardization strategies that facilitate consistent learning across heterogeneous platforms for a lightweight robot transformer.

\begin{figure*}[t]
  \centering
  \begin{minipage}[t]{0.62\textwidth}
    \centering
    \includegraphics[width=\linewidth]{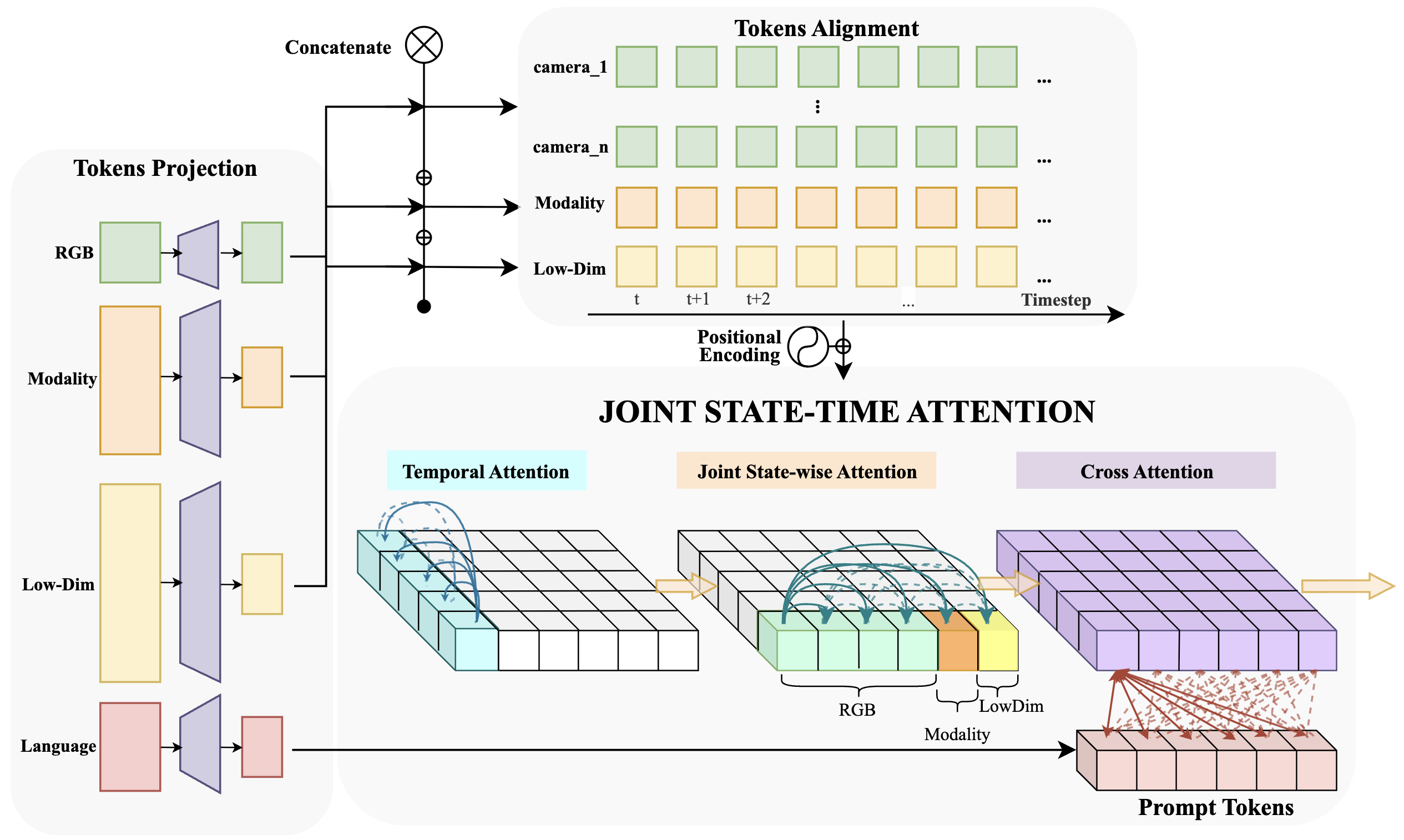}
    \caption*{(a)}
    \label{fig:encoder}
  \end{minipage}%
  \hspace{0.015\textwidth}
  \begin{minipage}[t]{0.27\textwidth}
    \centering
    \includegraphics[width=\linewidth]{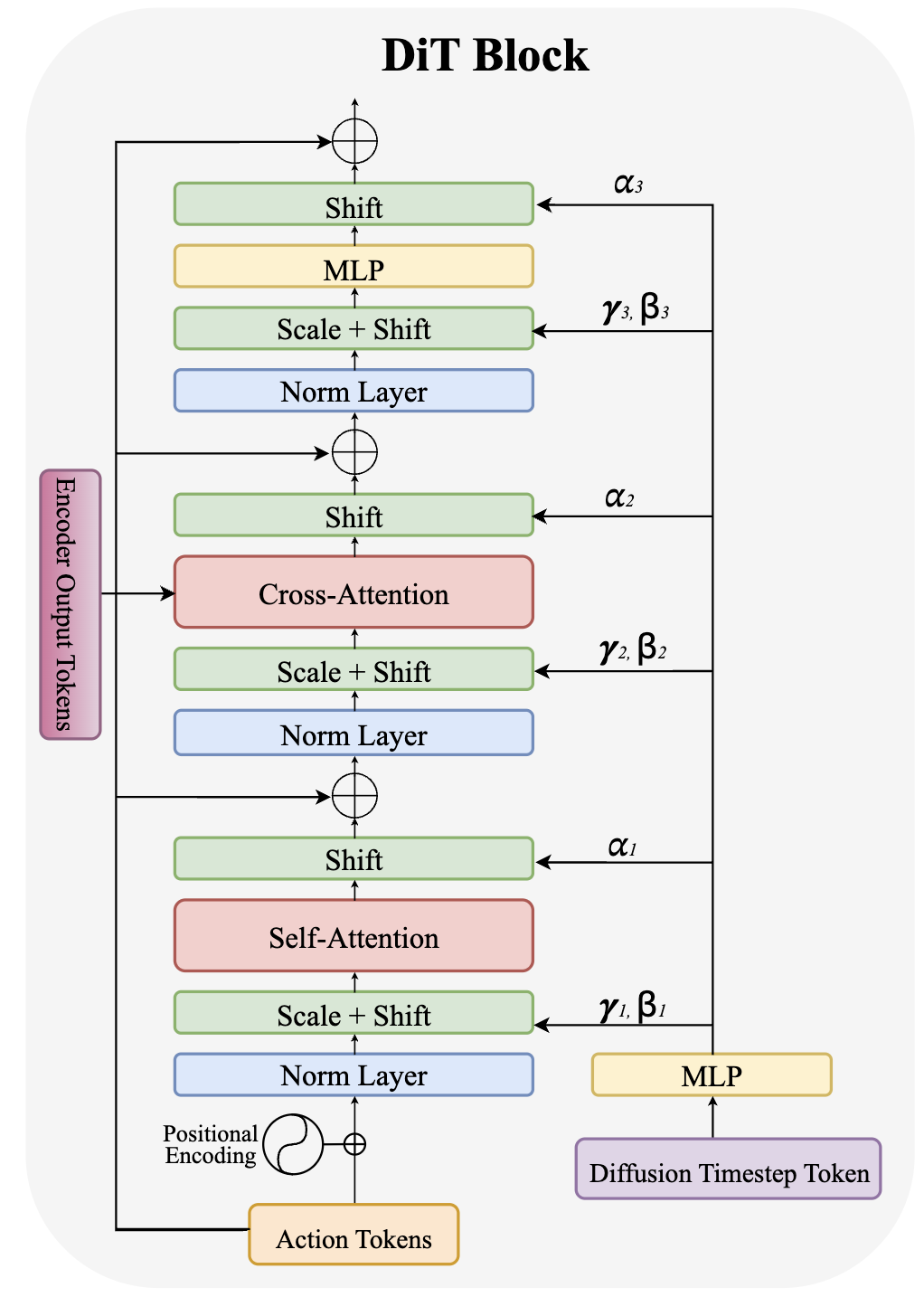}
    \caption*{(b)}
    \label{fig:decoder}
  \end{minipage}
  \vspace{4pt}  
  \caption{Overview of the Tenma encoder–decoder architecture. (a) illustrates the encoding pipeline, where cross-embodiment observations are projected into tokens, temporally aligned, and separated with a \textbf{modality token} to encode structural cues and handle variable token lengths across embodiments. These tokens are processed by \textbf{Joint State–Time Attention} for multimodal, temporal feature extraction and conditioned with language tokens. (b) depicts the \textbf{DiT block}, where noised action tokens are adaptively normalized via \textbf{AdaLN} using the diffusion timestep token and refined through \textbf{cross-attention} with encoded observation features.}
  \label{fig:tenma-arch}
\end{figure*}

\textbf{Transformer-based robot policies.} Transformer architectures are now central to imitation learning policies, with models such as ACT \cite{act}, RT-1 \cite{rt-1}, RT-2 \cite{rt-2}, and OpenVLA \cite{openvla} demonstrating the potential of sequence modeling for action generation. More recently, diffusion- and flow-based models \cite{svp, robograsp, srgrasp, dp, rdt, dit-policy, pi0, octo} have improved policy robustness by enabling diverse action sampling over complex observation spaces. However, these architectures often rely on either collapsing multi-modal features into a single token or applying full-sequence self-attention that can induce information bottlenecks and poor cross-modal integration. Moreover, most lightweight transformer policies continue to rely on ResNet backbones \cite{resnet} that lack the spatial capacity of modern vision transformers \cite{dp, act, octo, dit-policy}. While pretrained ViTs such as DINO \cite{dino} and CLIP \cite{clip} have shown strong generalization in other domains, their use in lightweight robotics models remains limited. In this work, we propose new observation feature extraction techniques that exploit the DINOv2 \cite{dinov2} vision and T5 \cite{t5} language features for richer feature extraction. What's more, we provide new observation learning and action diffusion conditioning techniques that outperform previous works by learning more robust, expressive features while maintaining high inference speeds.

\section{Problem Formulation}
The goal of our diffusion-based policy is to generate a sequence of goal-directed robot joint/Cartesian state actions conditioned on both natural language instructions and recent multimodal sensor observations. This formulation supports generalization across diverse robot embodiments and tasks by unifying perception, language understanding, and action synthesis within a single generative framework.

Specifically, let $\mathcal{L}$ denote the space of language commands, $\mathcal{X}$ the space of RGB video observations, and $\mathcal{Z}$ the space of proprioceptive states. At each time step $t \in \mathbb{N}^+$, the input to the policy is defined as:
\[
(\ell, o_t) \in \mathcal{L} \times \left(\mathcal{X}^h \times \mathcal{Z}^h\right),
\]
where $\ell$ is a natural language instruction, and $o_t = (X_{t-h+1:t},\ z_{t-h+1:t})$ represents an $h$-step observation window with $X_{t-h+1:t} \in \mathcal{X}^h$ denoting RGB frames and $z_{t-h+1:t} \in \mathcal{Z}^h$ denoting proprioceptive states.

The policy $\pi$ is defined as a function:
\[
\pi: \mathcal{L} \times \mathcal{X}^h \times \mathcal{Z}^h \rightarrow \mathcal{A}^n,
\]
where $\mathcal{A}$ is the action space and $n$ is the prediction horizon. The output sequence $a_{t:t+n-1} = (a_t, a_{t+1}, \ldots, a_{t+n-1})$ consists of future actions aimed at fulfilling the task encoded by $\ell$.

\subsection{Training Objective}
We instantiate $\pi$ as a conditional Denoising Diffusion Probabilistic Model (DDPM)~\cite{ddpm}, a class of generative models that synthesize outputs by reversing a gradual noising process. At each denoising step $k$, a noisy action trajectory $a_k$ is updated via:
\[
a_{k-1} = \alpha \left( a_k - \gamma \, \epsilon_\theta(a_k, k, o_t, \ell) \right) + \mathcal{N}(0, \sigma^2 I),
\]
where $\epsilon_\theta$ is the noise prediction network, and $\alpha$, $\gamma$, and $\sigma$ are parameters from a predefined diffusion schedule. The model is trained to minimize the discrepancy between the true noise $\epsilon_k$ and the predicted noise:
\[
\mathcal{L}_{\text{diffusion}} = \left\| \epsilon_k - \epsilon_\theta(a_t + \epsilon_k,\ k,\ o_t,\ \ell) \right\|^2_2,
\]
with $\epsilon_k \sim \mathcal{N}(0, I)$. Training is performed with $k = 1000$ diffusion steps using a cosine noise schedule~\cite{iddpm}, while inference is accelerated using deterministic sampling~\cite{ddim} with $k = 7$ steps.

\section{Tenma Model}
\textbf{Tenma} is a ViT-based \cite{vit} diffusion model designed for cross-embodiment robotic arm control. The architecture adopts an encoder–decoder structure: a Joint State–Time Encoder fuses multimodal observations with language prompts, while a diffusion decoder iteratively denoises future action sequences. To enable multitask learning across diverse embodiments, Tenma employs embodiment-normalization strategies that map heterogeneous joint configurations, camera viewpoints, and sensor layouts into a unified representation. The model is parameter-efficient, pairing a T5-small language encoder with ViT-S-sized transformer blocks for both encoder and decoder.  For stability, we employ RMSNorm \cite{rmsnorm} and QKNorm \cite{qknorm} throughout all non-frozen modules of the network.

\subsection{Proprioceptive \& Action Data Standardization}
We use a single \textbf{cross-embodiment standardizer}: map each robot state into a fixed-length table of canonical slots (e.g., joint angles/velocities, gripper, end-effector pose) with a binary mask for active entries, then apply per-embodiment min–max scaling to $[-1,1]$, with statistics kept in a dynamic, hashable store and inverted at deployment. This yields a constant-size, embodiment-agnostic token that stabilizes training across robots while avoiding the linear memory growth of per-embodiment projectors~\cite{gr00t, pi0}.

\subsection{Input Tokenization, Encoding \& Alignment}
Tenma converts each input stream into a sequence of tokens using modality-specific tokenizers (Table \ref{tab:tokenizers}). Each tokenizer maps inputs to a token sequence of shared latent space using modality-specific, two-layer MLPs with RMSNorm \cite{rmsnorm} and GELU \cite{gelu}. All tokens are projected to a fixed embedding dimension of 384, ensuring compatibility for transformer-based processing in subsequent modules.

\begin{table}[h]
\centering
\caption{Modality-specific tokenization methods in Tenma.}
\label{tab:tokenizers}
\renewcommand{\arraystretch}{1.15} 
\begin{tabularx}{\linewidth}{l p{0.20\linewidth} X}
\toprule
\textbf{Modality} & \textbf{Method} & \textbf{Details} \\
\midrule
\textbf{RGB} (multiview) & DINOv2-S \cite{dinov2} & Retains full patch sequence (no pooling) for robust, object-centric cues \cite{uni-vit}. \\
\textbf{Language} & T5-S \cite{t5} & Produces contextualized prompt tokens for conditioning. \\
\textbf{Proprioception} & Standardization & Aligns joint angles, velocities, and gripper states into a unified token with set positions, with a positional mask indicating which features are active at each timestep. \\
\bottomrule
\end{tabularx}
\end{table}

Temporal and spatial positions are encoded using sinusoidal embeddings \cite{attention}. Observation data (RGB and proprioceptive sensor data) is temporally aligned to form a unified sequence for transformer processing (Figure~\ref{fig:tenma-arch}). To handle variable camera counts across embodiments, a learnable modality token is inserted after each RGB block per timestep, marking modality boundaries for length-invariant fusion without padding. The resulting token sequence serves as the conditioning input for the diffusion process, guiding the denoising of action tokens.

\subsection{Observation Understanding}
Observation understanding occurs by performing self-attention and language conditioning on the tokenized observation data. This is performed in the Joint State-Time-Encoder (Figure \ref{fig:main}). This module utilizes our Joint State-Time Attention mechanism to factorize attention across time, state, and language to improve efficiency and stability when processing multimodal, cross-embodiment inputs. Each block applies:  
(1) \textbf{Temporal attention} along the time axis with a causal mask to preserve temporal ordering, enabling motion modeling without the quadratic cost of full attention.  
(2) \textbf{Joint state attention} within each timestep, where observation tokens perform joint self-attention. To support \emph{flexible cross-embodiment fusion} with variable RGB token sequence lengths, the \emph{modality token} signals the joint position between the two modalities. This attention is bidirectional, as no temporal constraints exist within a state.  
(3) \textbf{Cross-attention with language tokens}, applied bidirectionally to condition observation features on task prompts, guiding the attention to extract task-oriented features in the observation data.

Full attention scales quadratically with sequence length:
\[
\text{Full Attention: } O\big((TS + L)^2\big),
\]
where $T$, $S$, and $L$ denote timesteps, tokens per state, and language tokens, respectively. Our factorized design reduces this to:
\[
\text{Ours: } O(T^2 S + S^2 T + TS L),
\]
eliminating the $O(T^2 S^2)$ term and retaining efficient conditioning since $L \ll TS$. This factorization substantially improves inference efficiency while preserving performance comparable to full attention. Inference speed comparisons are shown in Table \ref{tab:model-size}.

\subsection{Action Sampling with Diffusion}
Tenma adopts a DiT-style transformer~\cite{dit} adapted for robotic action denoising (Figure~\ref{fig:tenma-arch}). Each block conditions on (i) the encoded observation sequence via \textbf{cross-attention} and (ii) a diffusion-timestep embedding through \textbf{AdaLN-zero}-based shift-and-scale modulation. This design enables \emph{stable}, \emph{structure-preserving} fusion of multimodal information; allowing each modality to contribute proportionally during the denoising process.

Self-attention among action tokens is bidirectional, while cross-attention to observation tokens is applied within corresponding timesteps to maintain temporal consistency. The decoder terminates with a two-layer MLP mapping hidden states to joint commands, with AdaLN modulators zero-initialized for stability. Action trajectories are generated by iteratively refining Gaussian noise using DDPM-style sampling~\cite{ddpm,iddpm,ddim}.

\section{Tenma for Bi‑Manual Tasks}
For bi-manual arm control, we use an observation history of $h = 2$ timesteps $(o_{t-h+1:t})$ and predict an action horizon of $n = 16$ steps $(a_{t:t+n-1})$, enabling short-term context and forward planning within a single inference pass. For model training on bi-manual arm tasks, we adopt a two–stage curriculum: (1) large‑scale cross‑embodiment pretraining, and (2) task‑specific fine‑tuning on a dual‑arm setup for food‑preparation. All training was performed on 8 A100 GPUs.

\paragraph{Cross-Embodiment Pretraining}
From the \textbf{X-Embodiment} corpus \cite{oxe} we select $\sim$2k demonstrations spanning 20 tabletop-manipulation tasks and 7 distinct robot embodiments. These include both single-arm and dual-arm systems, ensuring coverage across a wide spectrum of control configurations. Clips are set to 10Hz if timestamps exist, and left at native rate otherwise. This diverse slice supplies rich object variation, bi-manual behaviours, and broad embodiment coverage, equipping Tenma with strong low-level priors for cross-embodiment and bi-manual arm control before fine-tuning.

\paragraph{Domain Fine‑tuning}
We fine-tune on $\sim$450 human demonstrations collected with this setup, focusing on long-horizon food-prep tasks. Demonstrations deliberately vary objects, layouts, and lighting, and include perturbations and partial failures to encourage robustness. Each trajectory records (1) synchronized 4-view RGB at 10Hz, (2) end-effector state/action vectors, and (3) ten paraphrased language prompts, yielding multimodal data for adaptation.

\begin{table}[h]
\centering
\caption{Training hyperparameters.}
\label{tab:hyperparams}
\renewcommand{\arraystretch}{1.2}
\begin{tabular}{l c}
\toprule
\textbf{Hyperparameter} & \textbf{Value} \\
\midrule
Dropout                & 0.0 \\
Weight decay           & 0.0 \\
Gradient clipping      & None \\
Learning rate & $1 \times 10^{-4}$ \\
Batch size             & 32 \\
Global batch size      & 256 \\
Warmup steps           & 500 \\
Precision              & BF16 \\
Optimizer              & AdamW ($\beta_1=0.9$, $\beta_2=0.999$, $\epsilon=10^{-8}$) \\
EMA                    & Yes ($\mu = 0.9999$) \\
\bottomrule
\end{tabular}
\end{table}

\section{Experiments}

\subsection{Experiment Setup}

We evaluate policy performance across a range of generalization and robustness challenges, focusing on adaptation to generalizability of objects, scenes, and prompts. All experiments follow a fixed evaluation protocol with consistent checkpoints. To ensure realism, we focus on a physical kitchen-inspired setup rather than simulation environment. Because food-preparation tasks involve rich object properties, textures, and contact dynamics, keeping the evaluation grounded in the real world is essential for faithfully testing policy robustness.

We designed four tasks, each targeting a distinct aspect of manipulation:
\begin{itemize}
    \item \textbf{Pour}: Requires precise arm rotation and timing, sensitive to container variation.  
    \item \textbf{PickCup}: Tests grasp robustness across cups of different shapes and appearances.  
    \item \textbf{PlacePan}: Evaluates spatial reasoning through accurate pan placement on a rack.  
    \item \textbf{Stirring}: Challenges long-horizon control, from grasping a thin spoon to sustaining circular motions.  
\end{itemize}

Together, these tasks form a coherent and semantically meaningful domain for studying policy generalization and robustness. Each task and evaluation axis is demonstrated in Figure~\ref{fig:overview}.

\subsection{Hardware Setup}
Our experiments use a bi-manual robotic platform composed of two FAIRNO FR5 arms, supported by four Hikvision DS-E12A RGB cameras. The cameras provide complementary viewpoints, with two fixed overhead (high, low) and two wrist-mounted (left, right), offering both global workspace coverage and fine-grained object perspectives.

\subsection{Metrics and Evaluation Axes}

We report \textbf{Success Rate (SR)} as the primary evaluation metric, defined as the proportion of episodes in which the final task objective is achieved. SR provides a clear and interpretable measure of policy reliability, and is consistent with standard practice in prior work.  

To assess Tenma’s generalization and robustness, we evaluate SR along three axes that reflect the key distribution shifts in our experimental setup (Table~\ref{tab:evaluation-axes}). These axes capture performance in the nominal setting as well as under object and scene variations.  

\begin{table}[h]
\centering
\caption{Evaluation axes for assessing generalization and robustness.}
\label{tab:evaluation-axes}
\resizebox{\linewidth}{!}{
\begin{tabular}{p{0.33\linewidth} p{0.67\linewidth}}
\toprule
\textbf{Axis} & \textbf{Description} \\
\midrule
\textbf{1. In-Distribution} & Standard evaluation on tasks drawn directly from the fine-tuning distribution. \\
\textbf{2. Object-Shift} & Replace familiar objects with novel instances (e.g., different cups, pans, or spoons) to test intra-task generalization. \\
\textbf{3. Scene-Shift} & Vary background layouts, lighting conditions, and table configurations to measure spatial robustness. \\
\bottomrule
\end{tabular}}
\end{table}

\subsection{Model Variants and Ablation Study}

We evaluate Tenma against three representative diffusion–transformer policies of similar capacity, chosen for strong reported performance and architectural proximity so that comparisons emphasize design choices rather than model size. All comparative models were trained with the hyperparameters provided by the authors' open source implementations. We exclude very large foundation models ($>1$B parameters)~\cite{pi0,openvla,rt-2,rdt}, where scale effects dominate architectural contributions.

\newcolumntype{Y}{>{\centering\arraybackslash}X}

\begin{table}[h]
\centering
\caption{Model size and inference control frequency for the policies compared in the ablation. Parameter counts are approximate; frequencies were measured empirically using the authors’ open-source implementations under a common evaluation harness. Higher values (Hz) indicate faster inference.}
\label{tab:model-size}
\renewcommand{\arraystretch}{1.15}
\begin{tabularx}{\linewidth}{@{} l YY @{}}
\toprule
\textbf{Policy} & \textbf{Parameters} & \textbf{Freq.\ (Hz) $\uparrow$} \\
\midrule
\textbf{DiT-Policy}~\cite{dit-policy} & $\sim$90M & 31.0 \\
\textbf{Diffusion Policy}~\cite{dp}   & $\sim$50M  & 13.9 \\
\textbf{Octo}~\cite{octo}             & $\sim$200M & 10.6 \\
\textbf{Tenma}                 & $\sim$110M & 20.2 \\
\bottomrule
\end{tabularx}
\end{table}

Tables~\ref{tab:model-size} and~\ref{tab:model-features} summarize parameter counts, runtime frequency, and key design features. Parameter sizes are broadly comparable across methods. Tenma reaches \textbf{$\sim$ 20\,Hz} with 4 cameras on a RTX 3090 GPU, exceeding \emph{Diffusion Policy} and \emph{Octo} but trailing \emph{DiT-Policy} (31\,Hz).

\begin{table*}[h]
\centering
\caption{Comparison of Success Rate (SR, \%) across baseline models, tasks, and evaluation axes.}
\label{tab:results-comparison}
\resizebox{\textwidth}{!}{
\begin{tabular}{l|cccc|cccc|cccc}
\toprule
\multirow{2}{*}{\textbf{Task}}
  & \multicolumn{4}{c|}{\textbf{In-Distribution}}
  & \multicolumn{4}{c|}{\textbf{Object-Shift}}
  & \multicolumn{4}{c}{\textbf{Scene-Shift}} \\
\cmidrule(lr){2-5}\cmidrule(lr){6-9}\cmidrule(lr){10-13}
 & \textbf{Tenma} & \textbf{DiT-Policy} \cite{dit-policy} & \textbf{Diffusion Policy} \cite{dp}& \textbf{Octo} \cite{octo}
 & \textbf{Tenma} & \textbf{DiT-Policy} & \textbf{Diffusion Policy} & \textbf{Octo}
 & \textbf{Tenma} & \textbf{DiT-Policy} & \textbf{Diffusion Policy} & \textbf{Octo} \\
\midrule
Pour        & 93.75 & 14.38 & 12.30 & 10.27 & 61.54 & 5.49 & 4.40 & 3.11 & 88.10 & 6.47 & 5.10 & 5.25 \\
PickCup     & 100.00 & 24.25 & 17.00 & 13.50 & 87.50 & 7.23 & 8.32 & 5.52 & 100.00 & 5.16 & 4.60 & 7.91 \\
PlacePan    & 78.38 & 13.79 & 11.80 &  12.88 & 71.20 & 4.68 & 4.29 & 3.94 & 70.40 & 4.33 & 3.90 & 3.70 \\
Stirring    & 83.67 & 20.04 & 23.90 & 14.00 & 70.00 & 6.98 & 9.53 & 4.36 & 66.00 & 3.20 & 2.90 & 5.48 \\
\midrule
\textbf{Average} & \textbf{88.95} & 18.12 & 16.25 & 11.91 & \textbf{72.56} & 6.09 & 6.64 & 4.23 & \textbf{81.13} & 4.79 & 4.13 & 5.59 \\
\bottomrule
\end{tabular}}
\end{table*}

\begin{table}[h]
\centering
\caption{Architectural features of compared policies ($n$ = \# an arbitrary number of observation tokens). \emph{DiT-like} means very similar to the formal DiT architecture but with some small technical differences. \# Cond. Tokens are the number of tokens that the action tokens are conditioned on.}
\label{tab:model-features}
\renewcommand{\arraystretch}{1.15}
\begin{tabularx}{\linewidth}{@{} l l c c @{}}
\toprule
\textbf{Policy} & \textbf{Vision Encoder} & \textbf{Decoder} & \textbf{\# Cond. Tokens} \\
\midrule
\textbf{DiT-Policy}~\cite{dit-policy}       & ResNet-18 \cite{resnet}  & DiT \cite{dit} & 1 \\
\textbf{Diffusion Policy}~\cite{dp}         & ResNet-18  & \emph{DiT-like} & $1$/timestep \\
\textbf{Octo}~\cite{octo}                   & ResNet-50  & MLP & $\sim n$/timestep \\
\textbf{Tenma}                       & DINOv2-S~\cite{dinov2} & DiT & $n$/timestep \\
\bottomrule
\end{tabularx}
\end{table}

As demonstrated in Table \ref{tab:model-features}, this runtime profile is consistent with architectural differences: Tenma is the only model using a ViT-based encoder (DINOv2-S) rather than lightweight ResNet backbones, and Tenma (like Octo) processes \emph{$n$ observation tokens per timestep}, whereas DiT-Policy and Diffusion Policy compress observations to a \emph{single} token, reducing attention cost at potential information loss. Moreover, Octo is not a \emph{DiT-like} policy where it employs a diffusion transformer for denoising, rather it employs a compact ResNet–MLP diffusion head, trading latency for accuracy in our experiments as each model will decode actions an arbitrary number of times. Taken together, these factors indicate that Tenma’s $\sim$20\,Hz inference frequency reflects a favorable speed–accuracy trade-off given its richer encoder and multi-token representation.

\section{Results}
Table~\ref{tab:results-comparison} reports Success Rate (SR) across four tasks and three axes (in-distribution, object-shift, scene-shift). \textbf{Overall}, Tenma attains an average SR of $88.95\%$ in-distribution and $72.56\%$ / $81.13\%$ under object-/scene-shift, respectively.

\textbf{Task-specific observations.}
\emph{Pour} requires a multi-phase sequence (grasp $\to$ transport $\to$ tilt/hold $\to$ pause) with fine control of wrist rotation; object-shift introduces new container geometries (lip/diameter/mass distribution) that alter required tilt profiles, explaining the larger drop ($93.75\%\!\to\!61.54\%$). Under scene-shift, changes in camera pose and target placement are less detrimental ($88.10\%$), suggesting visual grounding is robust once a stable tilt strategy is learned.
\emph{PickCup} is a short-horizon grasp-and-lift with moderate pose variation; Tenma maintains near-ceiling SR (ID $100\%$, object $87.50\%$, scene $100\%$), indicating reliable grasp affordances and approach-vector selection even with texture/appearance changes.
\emph{PlacePan} involves precise placement onto a target region; errors arise from small pose misalignments and occlusions near the goal. Performance is moderate and relatively uniform across shifts (ID $78.38\%$, object $71.20\%$, scene $70.40\%$), consistent with sensitivity to final alignment rather than object identity.
\emph{Stirring} is long-horizon and contact-sensitive (maintaining circular trajectories, bowl depth, and grip on a thin handle); it degrades most under shifts (ID $83.67\%$, object $70.00\%$, scene $66.00\%$), reflecting accumulation of small pose errors and difficulty sustaining stable, periodic motion.

\begin{figure}[!t]
  \centering
  \includegraphics[width=\linewidth]{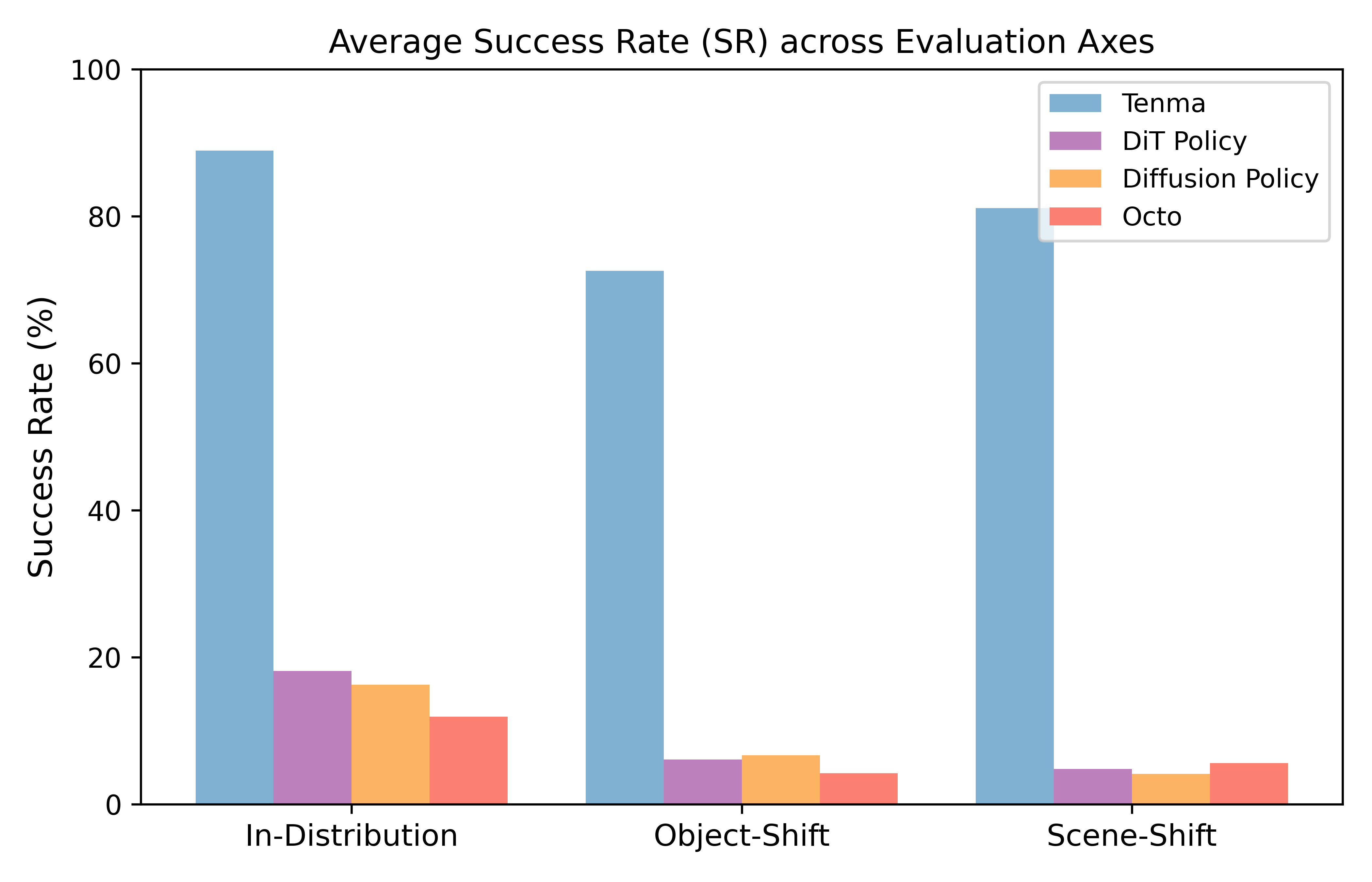}
  \caption{Visualization of average Success Rate (SR, \%) for Tenma and baseline models across three evaluation axes. This figure summarizes the quantitative results reported in Table~\ref{tab:results-comparison}, showing that Tenma consistently outperforms all baselines and maintains high performance under distribution shifts.
}
  \label{fig:results}
\end{figure}

\textbf{Baselines.} DiT-Policy averages $16.87\%$ in-distribution; rollouts frequently reach the object but show hesitation and unstable closures, leading to failures on \emph{PickCup}/\emph{Stirring} and mis-targeting on \emph{Pour}/\emph{PlacePan}. Octo averages $11.91\%$ and often exhibits early task confusion or stalled trajectories, with poor adaptation under both shifts. Diffusion Policy averages $14.25\%$; executions are steadier than Octo’s but commonly end in partial completions (drift, grasp loss), especially on \emph{Pour}/\emph{PlacePan}. These behaviors align with architectural constraints (Table~\ref{tab:model-features}): single-token conditioning or shallow decoders limit information flow needed for precise tilt control, final alignment, and sustained periodic motion.

\section{Discussion}
The results indicate that Tenma attains high in-distribution success (avg.\ SR $=88.95\%$) and degrades under distribution shifts (object-shift $72.56\%$, scene-shift $81.13\%$), consistent with the increased difficulty of novel objects and layouts (Table~\ref{tab:results-comparison}). In our evaluation, Tenma outperforms the state-of-the-art baselines (Octo, DiT, Diffusion Policy) across tasks and axes.

Dataset construction likely contributes to these outcomes. For pretraining, we curated a subset of Open X-Embodiment focused on tabletop/food-preparation tasks to ensure diversity of embodiments while maintaining domain coherence. For fine-tuning, we used targeted bi-manual robot demonstrations with an intentional but imbalanced allocation (e.g., \emph{PlacePan} has 50 episodes versus 100 for other tasks), which plausibly explains its lower in-distribution SR. The overall results are achieved with moderate data scale and a compact policy.

Architectural factors likely account for part of the performance gap. Tenma uses pretrained DINOv2-S (vision), providing stronger perceptual features than the ResNet-based baselines. A cross-embodiment standardizer yields a unified representation that supports stable training across heterogeneous robots, while the Joint State–Time encoder paired with a DiT-based decoder enables fine-grained, temporally coherent action synthesis. By contrast, baselines that employ shallow decoders or compress observations to a single conditioning token (Table~\ref{tab:model-features}) introduce information bottlenecks, which is consistent with their observed errors.

Despite using fewer demonstrations than DiT-Policy and Octo and, in Octo’s case, a smaller parameter budget, Tenma achieves higher success rates across benchmarks (Tables~\ref{tab:model-size}–\ref{tab:results-comparison}). Within our evaluation, this suggests that careful multimodal representation design and targeted data curation can provide greater benefits for scaling of data and model size. Emphasizing learning quality through stronger encoders, cross-embodiment normalization, and multi-token conditioning appears complementary to scale and merits further study.

\section{Conclusion}

We presented Tenma, a diffusion-transformer policy that unifies action representation and multimodal conditioning for embodied manipulation. By combining multi-embodiment pretraining, visual–language prompting, and joint state–time attention, Tenma achieves stable and generalizable performance across diverse tabletop tasks. This work demonstrates that robust, generalizable embodied policies can be trained efficiently by filtering pretraining data for domain relevance, leveraging multi-embodiment diversity, and integrating architectural components that jointly capture temporal, spatial, and semantic structure, offering a practical path forward for scalable manipulation learning. 

In summary, Tenma achieves this level of performance with modest dataset sizes and model scale, underscoring the efficiency of our approach. Ultimately, Tenma represents a step toward scalable, robust, and adaptable robotic policies, bringing embodied AI closer to deployment in the real world.

\printbibliography

@inproceedings{mujoco,
  title={MuJoCo: A physics engine for model-based control},
  author={Todorov, Emanuel and Erez, Tom and Tassa, Yuval},
  booktitle={2012 IEEE/RSJ International Conference on Intelligent Robots and Systems},
  pages={5026--5033},
  year={2012},
  organization={IEEE},
  doi={10.1109/IROS.2012.6386109}
}

@misc{isaacgym,
      title={Isaac Gym: High Performance GPU-Based Physics Simulation For Robot Learning}, 
      author={Viktor Makoviychuk and Lukasz Wawrzyniak and Yunrong Guo and Michelle Lu and Kier Storey and Miles Macklin and David Hoeller and Nikita Rudin and Arthur Allshire and Ankur Handa and Gavriel State},
      year={2021},
      journal={arXiv preprint arXiv:2108.10470}
}

@misc{isaacsim,
  title = {NVIDIA Isaac Sim},
  howpublished = {\url{https://developer.nvidia.com/isaac/sim}},
  note = {Accessed: 2025-07-16},
  author = {NVIDIA},
  publisher = {NVIDIA}
}

@misc{gr00t,
      title={GR00T N1: An Open Foundation Model for Generalist Humanoid Robots}, 
      author={NVIDIA and : and Johan Bjorck and Fernando Castañeda and Nikita Cherniadev and Xingye Da and Runyu Ding and Linxi "Jim" Fan and Yu Fang and Dieter Fox and Fengyuan Hu and Spencer Huang and Joel Jang and Zhenyu Jiang and Jan Kautz and Kaushil Kundalia and Lawrence Lao and Zhiqi Li and Zongyu Lin and Kevin Lin and Guilin Liu and Edith Llontop and Loic Magne and Ajay Mandlekar and Avnish Narayan and Soroush Nasiriany and Scott Reed and You Liang Tan and Guanzhi Wang and Zu Wang and Jing Wang and Qi Wang and Jiannan Xiang and Yuqi Xie and Yinzhen Xu and Zhenjia Xu and Seonghyeon Ye and Zhiding Yu and Ao Zhang and Hao Zhang and Yizhou Zhao and Ruijie Zheng and Yuke Zhu},
      year={2025},
      eprint={2503.14734},
      archivePrefix={arXiv},
      primaryClass={cs.RO},
      url={https://arxiv.org/abs/2503.14734}, 
}

@misc{lbm,
      title={A Careful Examination of Large Behavior Models for Multitask Dexterous Manipulation}, 
      author={TRI LBM Team and Jose Barreiros and Andrew Beaulieu and Aditya Bhat and Rick Cory and Eric Cousineau and Hongkai Dai and Ching-Hsin Fang and Kunimatsu Hashimoto and Muhammad Zubair Irshad and Masha Itkina and Naveen Kuppuswamy and Kuan-Hui Lee and Katherine Liu and Dale McConachie and Ian McMahon and Haruki Nishimura and Calder Phillips-Grafflin and Charles Richter and Paarth Shah and Krishnan Srinivasan and Blake Wulfe and Chen Xu and Mengchao Zhang and Alex Alspach and Maya Angeles and Kushal Arora and Vitor Campagnolo Guizilini and Alejandro Castro and Dian Chen and Ting-Sheng Chu and Sam Creasey and Sean Curtis and Richard Denitto and Emma Dixon and Eric Dusel and Matthew Ferreira and Aimee Goncalves and Grant Gould and Damrong Guoy and Swati Gupta and Xuchen Han and Kyle Hatch and Brendan Hathaway and Allison Henry and Hillel Hochsztein and Phoebe Horgan and Shun Iwase and Donovon Jackson and Siddharth Karamcheti and Sedrick Keh and Joseph Masterjohn and Jean Mercat and Patrick Miller and Paul Mitiguy and Tony Nguyen and Jeremy Nimmer and Yuki Noguchi and Reko Ong and Aykut Onol and Owen Pfannenstiehl and Richard Poyner and Leticia Priebe Mendes Rocha and Gordon Richardson and Christopher Rodriguez and Derick Seale and Michael Sherman and Mariah Smith-Jones and David Tago and Pavel Tokmakov and Matthew Tran and Basile Van Hoorick and Igor Vasiljevic and Sergey Zakharov and Mark Zolotas and Rares Ambrus and Kerri Fetzer-Borelli and Benjamin Burchfiel and Hadas Kress-Gazit and Siyuan Feng and Stacie Ford and Russ Tedrake},
      year={2025},
      eprint={2507.05331},
      archivePrefix={arXiv},
      primaryClass={cs.RO},
      url={https://arxiv.org/abs/2507.05331}, 
}

@misc{droid,
      title={DROID: A Large-Scale In-The-Wild Robot Manipulation Dataset}, 
      author={Alexander Khazatsky and Karl Pertsch and Suraj Nair and Ashwin Balakrishna and Sudeep Dasari and Siddharth Karamcheti and Soroush Nasiriany and Mohan Kumar Srirama and Lawrence Yunliang Chen and Kirsty Ellis and Peter David Fagan and Joey Hejna and Masha Itkina and Marion Lepert and Yecheng Jason Ma and Patrick Tree Miller and Jimmy Wu and Suneel Belkhale and Shivin Dass and Huy Ha and Arhan Jain and Abraham Lee and Youngwoon Lee and Marius Memmel and Sungjae Park and Ilija Radosavovic and Kaiyuan Wang and Albert Zhan and Kevin Black and Cheng Chi and Kyle Beltran Hatch and Shan Lin and Jingpei Lu and Jean Mercat and Abdul Rehman and Pannag R Sanketi and Archit Sharma and Cody Simpson and Quan Vuong and Homer Rich Walke and Blake Wulfe and Ted Xiao and Jonathan Heewon Yang and Arefeh Yavary and Tony Z. Zhao and Christopher Agia and Rohan Baijal and Mateo Guaman Castro and Daphne Chen and Qiuyu Chen and Trinity Chung and Jaimyn Drake and Ethan Paul Foster and Jensen Gao and Vitor Guizilini and David Antonio Herrera and Minho Heo and Kyle Hsu and Jiaheng Hu and Muhammad Zubair Irshad and Donovon Jackson and Charlotte Le and Yunshuang Li and Kevin Lin and Roy Lin and Zehan Ma and Abhiram Maddukuri and Suvir Mirchandani and Daniel Morton and Tony Nguyen and Abigail O'Neill and Rosario Scalise and Derick Seale and Victor Son and Stephen Tian and Emi Tran and Andrew E. Wang and Yilin Wu and Annie Xie and Jingyun Yang and Patrick Yin and Yunchu Zhang and Osbert Bastani and Glen Berseth and Jeannette Bohg and Ken Goldberg and Abhinav Gupta and Abhishek Gupta and Dinesh Jayaraman and Joseph J Lim and Jitendra Malik and Roberto Martín-Martín and Subramanian Ramamoorthy and Dorsa Sadigh and Shuran Song and Jiajun Wu and Michael C. Yip and Yuke Zhu and Thomas Kollar and Sergey Levine and Chelsea Finn},
      year={2025},
      eprint={2403.12945},
      archivePrefix={arXiv},
      primaryClass={cs.RO},
      url={https://arxiv.org/abs/2403.12945}, 
}

@misc{oxe,
      title={Open X-Embodiment: Robotic Learning Datasets and RT-X Models}, 
      author={Embodiment Collaboration and Abby O'Neill and Abdul Rehman and Abhinav Gupta and Abhiram Maddukuri and Abhishek Gupta and Abhishek Padalkar and Abraham Lee and Acorn Pooley and Agrim Gupta and Ajay Mandlekar and Ajinkya Jain and Albert Tung and Alex Bewley and Alex Herzog and Alex Irpan and Alexander Khazatsky and Anant Rai and Anchit Gupta and Andrew Wang and Andrey Kolobov and Anikait Singh and Animesh Garg and Aniruddha Kembhavi and Annie Xie and Anthony Brohan and Antonin Raffin and Archit Sharma and Arefeh Yavary and Arhan Jain and Ashwin Balakrishna and Ayzaan Wahid and Ben Burgess-Limerick and Beomjoon Kim and Bernhard Schölkopf and Blake Wulfe and Brian Ichter and Cewu Lu and Charles Xu and Charlotte Le and Chelsea Finn and Chen Wang and Chenfeng Xu and Cheng Chi and Chenguang Huang and Christine Chan and Christopher Agia and Chuer Pan and Chuyuan Fu and Coline Devin and Danfei Xu and Daniel Morton and Danny Driess and Daphne Chen and Deepak Pathak and Dhruv Shah and Dieter Büchler and Dinesh Jayaraman and Dmitry Kalashnikov and Dorsa Sadigh and Edward Johns and Ethan Foster and Fangchen Liu and Federico Ceola and Fei Xia and Feiyu Zhao and Felipe Vieira Frujeri and Freek Stulp and Gaoyue Zhou and Gaurav S. Sukhatme and Gautam Salhotra and Ge Yan and Gilbert Feng and Giulio Schiavi and Glen Berseth and Gregory Kahn and Guangwen Yang and Guanzhi Wang and Hao Su and Hao-Shu Fang and Haochen Shi and Henghui Bao and Heni Ben Amor and Henrik I Christensen and Hiroki Furuta and Homanga Bharadhwaj and Homer Walke and Hongjie Fang and Huy Ha and Igor Mordatch and Ilija Radosavovic and Isabel Leal and Jacky Liang and Jad Abou-Chakra and Jaehyung Kim and Jaimyn Drake and Jan Peters and Jan Schneider and Jasmine Hsu and Jay Vakil and Jeannette Bohg and Jeffrey Bingham and Jeffrey Wu and Jensen Gao and Jiaheng Hu and Jiajun Wu and Jialin Wu and Jiankai Sun and Jianlan Luo and Jiayuan Gu and Jie Tan and Jihoon Oh and Jimmy Wu and Jingpei Lu and Jingyun Yang and Jitendra Malik and João Silvério and Joey Hejna and Jonathan Booher and Jonathan Tompson and Jonathan Yang and Jordi Salvador and Joseph J. Lim and Junhyek Han and Kaiyuan Wang and Kanishka Rao and Karl Pertsch and Karol Hausman and Keegan Go and Keerthana Gopalakrishnan and Ken Goldberg and Kendra Byrne and Kenneth Oslund and Kento Kawaharazuka and Kevin Black and Kevin Lin and Kevin Zhang and Kiana Ehsani and Kiran Lekkala and Kirsty Ellis and Krishan Rana and Krishnan Srinivasan and Kuan Fang and Kunal Pratap Singh and Kuo-Hao Zeng and Kyle Hatch and Kyle Hsu and Laurent Itti and Lawrence Yunliang Chen and Lerrel Pinto and Li Fei-Fei and Liam Tan and Linxi "Jim" Fan and Lionel Ott and Lisa Lee and Luca Weihs and Magnum Chen and Marion Lepert and Marius Memmel and Masayoshi Tomizuka and Masha Itkina and Mateo Guaman Castro and Max Spero and Maximilian Du and Michael Ahn and Michael C. Yip and Mingtong Zhang and Mingyu Ding and Minho Heo and Mohan Kumar Srirama and Mohit Sharma and Moo Jin Kim and Muhammad Zubair Irshad and Naoaki Kanazawa and Nicklas Hansen and Nicolas Heess and Nikhil J Joshi and Niko Suenderhauf and Ning Liu and Norman Di Palo and Nur Muhammad Mahi Shafiullah and Oier Mees and Oliver Kroemer and Osbert Bastani and Pannag R Sanketi and Patrick "Tree" Miller and Patrick Yin and Paul Wohlhart and Peng Xu and Peter David Fagan and Peter Mitrano and Pierre Sermanet and Pieter Abbeel and Priya Sundaresan and Qiuyu Chen and Quan Vuong and Rafael Rafailov and Ran Tian and Ria Doshi and Roberto Martín-Martín and Rohan Baijal and Rosario Scalise and Rose Hendrix and Roy Lin and Runjia Qian and Ruohan Zhang and Russell Mendonca and Rutav Shah and Ryan Hoque and Ryan Julian and Samuel Bustamante and Sean Kirmani and Sergey Levine and Shan Lin and Sherry Moore and Shikhar Bahl and Shivin Dass and Shubham Sonawani and Shubham Tulsiani and Shuran Song and Sichun Xu and Siddhant Haldar and Siddharth Karamcheti and Simeon Adebola and Simon Guist and Soroush Nasiriany and Stefan Schaal and Stefan Welker and Stephen Tian and Subramanian Ramamoorthy and Sudeep Dasari and Suneel Belkhale and Sungjae Park and Suraj Nair and Suvir Mirchandani and Takayuki Osa and Tanmay Gupta and Tatsuya Harada and Tatsuya Matsushima and Ted Xiao and Thomas Kollar and Tianhe Yu and Tianli Ding and Todor Davchev and Tony Z. Zhao and Travis Armstrong and Trevor Darrell and Trinity Chung and Vidhi Jain and Vikash Kumar and Vincent Vanhoucke and Vitor Guizilini and Wei Zhan and Wenxuan Zhou and Wolfram Burgard and Xi Chen and Xiangyu Chen and Xiaolong Wang and Xinghao Zhu and Xinyang Geng and Xiyuan Liu and Xu Liangwei and Xuanlin Li and Yansong Pang and Yao Lu and Yecheng Jason Ma and Yejin Kim and Yevgen Chebotar and Yifan Zhou and Yifeng Zhu and Yilin Wu and Ying Xu and Yixuan Wang and Yonatan Bisk and Yongqiang Dou and Yoonyoung Cho and Youngwoon Lee and Yuchen Cui and Yue Cao and Yueh-Hua Wu and Yujin Tang and Yuke Zhu and Yunchu Zhang and Yunfan Jiang and Yunshuang Li and Yunzhu Li and Yusuke Iwasawa and Yutaka Matsuo and Zehan Ma and Zhuo Xu and Zichen Jeff Cui and Zichen Zhang and Zipeng Fu and Zipeng Lin},
      year={2025},
      eprint={2310.08864},
      archivePrefix={arXiv},
      primaryClass={cs.RO},
      url={https://arxiv.org/abs/2310.08864}, 
}

@misc{agibot,
      title={AgiBot World Colosseo: A Large-scale Manipulation Platform for Scalable and Intelligent Embodied Systems}, 
      author={AgiBot-World-Contributors and Qingwen Bu and Jisong Cai and Li Chen and Xiuqi Cui and Yan Ding and Siyuan Feng and Shenyuan Gao and Xindong He and Xuan Hu and Xu Huang and Shu Jiang and Yuxin Jiang and Cheng Jing and Hongyang Li and Jialu Li and Chiming Liu and Yi Liu and Yuxiang Lu and Jianlan Luo and Ping Luo and Yao Mu and Yuehan Niu and Yixuan Pan and Jiangmiao Pang and Yu Qiao and Guanghui Ren and Cheng Ruan and Jiaqi Shan and Yongjian Shen and Chengshi Shi and Mingkang Shi and Modi Shi and Chonghao Sima and Jianheng Song and Huijie Wang and Wenhao Wang and Dafeng Wei and Chengen Xie and Guo Xu and Junchi Yan and Cunbiao Yang and Lei Yang and Shukai Yang and Maoqing Yao and Jia Zeng and Chi Zhang and Qinglin Zhang and Bin Zhao and Chengyue Zhao and Jiaqi Zhao and Jianchao Zhu},
      year={2025},
      eprint={2503.06669},
      archivePrefix={arXiv},
      primaryClass={cs.RO},
      url={https://arxiv.org/abs/2503.06669}, 
}

@misc{act,
      title={Learning Fine-Grained Bimanual Manipulation with Low-Cost Hardware}, 
      author={Tony Z. Zhao and Vikash Kumar and Sergey Levine and Chelsea Finn},
      year={2023},
      eprint={2304.13705},
      archivePrefix={arXiv},
      primaryClass={cs.RO},
      url={https://arxiv.org/abs/2304.13705}, 
}

@misc{rt-1,
      title={RT-1: Robotics Transformer for Real-World Control at Scale}, 
      author={Anthony Brohan and Noah Brown and Justice Carbajal and Yevgen Chebotar and Joseph Dabis and Chelsea Finn and Keerthana Gopalakrishnan and Karol Hausman and Alex Herzog and Jasmine Hsu and Julian Ibarz and Brian Ichter and Alex Irpan and Tomas Jackson and Sally Jesmonth and Nikhil J Joshi and Ryan Julian and Dmitry Kalashnikov and Yuheng Kuang and Isabel Leal and Kuang-Huei Lee and Sergey Levine and Yao Lu and Utsav Malla and Deeksha Manjunath and Igor Mordatch and Ofir Nachum and Carolina Parada and Jodilyn Peralta and Emily Perez and Karl Pertsch and Jornell Quiambao and Kanishka Rao and Michael Ryoo and Grecia Salazar and Pannag Sanketi and Kevin Sayed and Jaspiar Singh and Sumedh Sontakke and Austin Stone and Clayton Tan and Huong Tran and Vincent Vanhoucke and Steve Vega and Quan Vuong and Fei Xia and Ted Xiao and Peng Xu and Sichun Xu and Tianhe Yu and Brianna Zitkovich},
      year={2023},
      eprint={2212.06817},
      archivePrefix={arXiv},
      primaryClass={cs.RO},
      url={https://arxiv.org/abs/2212.06817}, 
}

@misc{rt-2,
      title={RT-2: Vision-Language-Action Models Transfer Web Knowledge to Robotic Control}, 
      author={Anthony Brohan and Noah Brown and Justice Carbajal and Yevgen Chebotar and Xi Chen and Krzysztof Choromanski and Tianli Ding and Danny Driess and Avinava Dubey and Chelsea Finn and Pete Florence and Chuyuan Fu and Montse Gonzalez Arenas and Keerthana Gopalakrishnan and Kehang Han and Karol Hausman and Alexander Herzog and Jasmine Hsu and Brian Ichter and Alex Irpan and Nikhil Joshi and Ryan Julian and Dmitry Kalashnikov and Yuheng Kuang and Isabel Leal and Lisa Lee and Tsang-Wei Edward Lee and Sergey Levine and Yao Lu and Henryk Michalewski and Igor Mordatch and Karl Pertsch and Kanishka Rao and Krista Reymann and Michael Ryoo and Grecia Salazar and Pannag Sanketi and Pierre Sermanet and Jaspiar Singh and Anikait Singh and Radu Soricut and Huong Tran and Vincent Vanhoucke and Quan Vuong and Ayzaan Wahid and Stefan Welker and Paul Wohlhart and Jialin Wu and Fei Xia and Ted Xiao and Peng Xu and Sichun Xu and Tianhe Yu and Brianna Zitkovich},
      year={2023},
      eprint={2307.15818},
      archivePrefix={arXiv},
      primaryClass={cs.RO},
      url={https://arxiv.org/abs/2307.15818}, 
}

@misc{openvla,
      title={OpenVLA: An Open-Source Vision-Language-Action Model}, 
      author={Moo Jin Kim and Karl Pertsch and Siddharth Karamcheti and Ted Xiao and Ashwin Balakrishna and Suraj Nair and Rafael Rafailov and Ethan Foster and Grace Lam and Pannag Sanketi and Quan Vuong and Thomas Kollar and Benjamin Burchfiel and Russ Tedrake and Dorsa Sadigh and Sergey Levine and Percy Liang and Chelsea Finn},
      year={2024},
      eprint={2406.09246},
      archivePrefix={arXiv},
      primaryClass={cs.RO},
      url={https://arxiv.org/abs/2406.09246}, 
}

@misc{dp,
      title={Diffusion Policy: Visuomotor Policy Learning via Action Diffusion}, 
      author={Cheng Chi and Zhenjia Xu and Siyuan Feng and Eric Cousineau and Yilun Du and Benjamin Burchfiel and Russ Tedrake and Shuran Song},
      year={2024},
      eprint={2303.04137},
      archivePrefix={arXiv},
      primaryClass={cs.RO},
      url={https://arxiv.org/abs/2303.04137}, 
}

@misc{rdt,
      title={RDT-1B: a Diffusion Foundation Model for Bimanual Manipulation}, 
      author={Songming Liu and Lingxuan Wu and Bangguo Li and Hengkai Tan and Huayu Chen and Zhengyi Wang and Ke Xu and Hang Su and Jun Zhu},
      year={2025},
      eprint={2410.07864},
      archivePrefix={arXiv},
      primaryClass={cs.RO},
      url={https://arxiv.org/abs/2410.07864}, 
}

@misc{dit-policy,
      title={The Ingredients for Robotic Diffusion Transformers}, 
      author={Sudeep Dasari and Oier Mees and Sebastian Zhao and Mohan Kumar Srirama and Sergey Levine},
      year={2024},
      eprint={2410.10088},
      archivePrefix={arXiv},
      primaryClass={cs.RO},
      url={https://arxiv.org/abs/2410.10088}, 
}

@misc{octo,
      title={Octo: An Open-Source Generalist Robot Policy}, 
      author={Octo Model Team and Dibya Ghosh and Homer Walke and Karl Pertsch and Kevin Black and Oier Mees and Sudeep Dasari and Joey Hejna and Tobias Kreiman and Charles Xu and Jianlan Luo and You Liang Tan and Lawrence Yunliang Chen and Pannag Sanketi and Quan Vuong and Ted Xiao and Dorsa Sadigh and Chelsea Finn and Sergey Levine},
      year={2024},
      eprint={2405.12213},
      archivePrefix={arXiv},
      primaryClass={cs.RO},
      url={https://arxiv.org/abs/2405.12213}, 
}

@misc{pi0,
      title={$\pi_0$: A Vision-Language-Action Flow Model for General Robot Control}, 
      author={Kevin Black and Noah Brown and Danny Driess and Adnan Esmail and Michael Equi and Chelsea Finn and Niccolo Fusai and Lachy Groom and Karol Hausman and Brian Ichter and Szymon Jakubczak and Tim Jones and Liyiming Ke and Sergey Levine and Adrian Li-Bell and Mohith Mothukuri and Suraj Nair and Karl Pertsch and Lucy Xiaoyang Shi and James Tanner and Quan Vuong and Anna Walling and Haohuan Wang and Ury Zhilinsky},
      year={2024},
      eprint={2410.24164},
      archivePrefix={arXiv},
      primaryClass={cs.LG},
      url={https://arxiv.org/abs/2410.24164}, 
}

@misc{resnet,
      title={Deep Residual Learning for Image Recognition}, 
      author={Kaiming He and Xiangyu Zhang and Shaoqing Ren and Jian Sun},
      year={2015},
      eprint={1512.03385},
      archivePrefix={arXiv},
      primaryClass={cs.CV},
      url={https://arxiv.org/abs/1512.03385}, 
}

@misc{vit,
      title={An Image is Worth 16x16 Words: Transformers for Image Recognition at Scale}, 
      author={Alexey Dosovitskiy and Lucas Beyer and Alexander Kolesnikov and Dirk Weissenborn and Xiaohua Zhai and Thomas Unterthiner and Mostafa Dehghani and Matthias Minderer and Georg Heigold and Sylvain Gelly and Jakob Uszkoreit and Neil Houlsby},
      year={2021},
      eprint={2010.11929},
      archivePrefix={arXiv},
      primaryClass={cs.CV},
      url={https://arxiv.org/abs/2010.11929}, 
}

@misc{dino,
      title={Emerging Properties in Self-Supervised Vision Transformers}, 
      author={Mathilde Caron and Hugo Touvron and Ishan Misra and Hervé Jégou and Julien Mairal and Piotr Bojanowski and Armand Joulin},
      year={2021},
      eprint={2104.14294},
      archivePrefix={arXiv},
      primaryClass={cs.CV},
      url={https://arxiv.org/abs/2104.14294}, 
}

@misc{clip,
      title={Learning Transferable Visual Models From Natural Language Supervision}, 
      author={Alec Radford and Jong Wook Kim and Chris Hallacy and Aditya Ramesh and Gabriel Goh and Sandhini Agarwal and Girish Sastry and Amanda Askell and Pamela Mishkin and Jack Clark and Gretchen Krueger and Ilya Sutskever},
      year={2021},
      eprint={2103.00020},
      archivePrefix={arXiv},
      primaryClass={cs.CV},
      url={https://arxiv.org/abs/2103.00020}, 
}

@misc{dinov2,
      title={DINOv2: Learning Robust Visual Features without Supervision}, 
      author={Maxime Oquab and Timothée Darcet and Théo Moutakanni and Huy Vo and Marc Szafraniec and Vasil Khalidov and Pierre Fernandez and Daniel Haziza and Francisco Massa and Alaaeldin El-Nouby and Mahmoud Assran and Nicolas Ballas and Wojciech Galuba and Russell Howes and Po-Yao Huang and Shang-Wen Li and Ishan Misra and Michael Rabbat and Vasu Sharma and Gabriel Synnaeve and Hu Xu and Hervé Jegou and Julien Mairal and Patrick Labatut and Armand Joulin and Piotr Bojanowski},
      year={2024},
      eprint={2304.07193},
      archivePrefix={arXiv},
      primaryClass={cs.CV},
      url={https://arxiv.org/abs/2304.07193}, 
}

@misc{t5,
      title={Exploring the Limits of Transfer Learning with a Unified Text-to-Text Transformer}, 
      author={Colin Raffel and Noam Shazeer and Adam Roberts and Katherine Lee and Sharan Narang and Michael Matena and Yanqi Zhou and Wei Li and Peter J. Liu},
      year={2023},
      eprint={1910.10683},
      archivePrefix={arXiv},
      primaryClass={cs.LG},
      url={https://arxiv.org/abs/1910.10683}, 
}

@misc{umi,
      title={Universal Manipulation Interface: In-The-Wild Robot Teaching Without In-The-Wild Robots}, 
      author={Cheng Chi and Zhenjia Xu and Chuer Pan and Eric Cousineau and Benjamin Burchfiel and Siyuan Feng and Russ Tedrake and Shuran Song},
      year={2024},
      eprint={2402.10329},
      archivePrefix={arXiv},
      primaryClass={cs.RO},
      url={https://arxiv.org/abs/2402.10329}, 
}

@misc{ddpm,
      title={Denoising Diffusion Probabilistic Models}, 
      author={Jonathan Ho and Ajay Jain and Pieter Abbeel},
      year={2020},
      eprint={2006.11239},
      archivePrefix={arXiv},
      primaryClass={cs.LG},
      url={https://arxiv.org/abs/2006.11239}, 
}

@misc{iddpm,
      title={Improved Denoising Diffusion Probabilistic Models}, 
      author={Alex Nichol and Prafulla Dhariwal},
      year={2021},
      eprint={2102.09672},
      archivePrefix={arXiv},
      primaryClass={cs.LG},
      url={https://arxiv.org/abs/2102.09672}, 
}

@misc{ddim,
      title={Denoising Diffusion Implicit Models}, 
      author={Jiaming Song and Chenlin Meng and Stefano Ermon},
      year={2022},
      eprint={2010.02502},
      archivePrefix={arXiv},
      primaryClass={cs.LG},
      url={https://arxiv.org/abs/2010.02502}, 
}

@misc{attention,
      title={Attention Is All You Need}, 
      author={Ashish Vaswani and Noam Shazeer and Niki Parmar and Jakob Uszkoreit and Llion Jones and Aidan N. Gomez and Lukasz Kaiser and Illia Polosukhin},
      year={2023},
      eprint={1706.03762},
      archivePrefix={arXiv},
      primaryClass={cs.CL},
      url={https://arxiv.org/abs/1706.03762}, 
}

@misc{uni-vit,
      title={Unifying Specialized Visual Encoders for Video Language Models}, 
      author={Jihoon Chung and Tyler Zhu and Max Gonzalez Saez-Diez and Juan Carlos Niebles and Honglu Zhou and Olga Russakovsky},
      year={2025},
      eprint={2501.01426},
      archivePrefix={arXiv},
      primaryClass={cs.CV},
      url={https://arxiv.org/abs/2501.01426}, 
}

@misc{dit,
      title={Scalable Diffusion Models with Transformers}, 
      author={William Peebles and Saining Xie},
      year={2023},
      eprint={2212.09748},
      archivePrefix={arXiv},
      primaryClass={cs.CV},
      url={https://arxiv.org/abs/2212.09748}, 
}

@misc{gelu,
      title={Gaussian Error Linear Units (GELUs)}, 
      author={Dan Hendrycks and Kevin Gimpel},
      year={2023},
      eprint={1606.08415},
      archivePrefix={arXiv},
      primaryClass={cs.LG},
      url={https://arxiv.org/abs/1606.08415}, 
}

@misc{qknorm,
      title={Query-Key Normalization for Transformers}, 
      author={Alex Henry and Prudhvi Raj Dachapally and Shubham Pawar and Yuxuan Chen},
      year={2020},
      eprint={2010.04245},
      archivePrefix={arXiv},
      primaryClass={cs.CL},
      url={https://arxiv.org/abs/2010.04245}, 
}

@misc{rmsnorm,
      title={Root Mean Square Layer Normalization}, 
      author={Biao Zhang and Rico Sennrich},
      year={2019},
      eprint={1910.07467},
      archivePrefix={arXiv},
      primaryClass={cs.LG},
      url={https://arxiv.org/abs/1910.07467}, 
}

@misc{gpt,
      title={Language Models are Few-Shot Learners}, 
      author={Tom B. Brown and Benjamin Mann and Nick Ryder and Melanie Subbiah and Jared Kaplan and Prafulla Dhariwal and Arvind Neelakantan and Pranav Shyam and Girish Sastry and Amanda Askell and Sandhini Agarwal and Ariel Herbert-Voss and Gretchen Krueger and Tom Henighan and Rewon Child and Aditya Ramesh and Daniel M. Ziegler and Jeffrey Wu and Clemens Winter and Christopher Hesse and Mark Chen and Eric Sigler and Mateusz Litwin and Scott Gray and Benjamin Chess and Jack Clark and Christopher Berner and Sam McCandlish and Alec Radford and Ilya Sutskever and Dario Amodei},
      year={2020},
      eprint={2005.14165},
      archivePrefix={arXiv},
      primaryClass={cs.CL},
      url={https://arxiv.org/abs/2005.14165}, 
}

@misc{llama,
      title={LLaMA: Open and Efficient Foundation Language Models}, 
      author={Hugo Touvron and Thibaut Lavril and Gautier Izacard and Xavier Martinet and Marie-Anne Lachaux and Timothée Lacroix and Baptiste Rozière and Naman Goyal and Eric Hambro and Faisal Azhar and Aurelien Rodriguez and Armand Joulin and Edouard Grave and Guillaume Lample},
      year={2023},
      eprint={2302.13971},
      archivePrefix={arXiv},
      primaryClass={cs.CL},
      url={https://arxiv.org/abs/2302.13971}, 
}

@misc{fp3,
      title={FP3: A 3D Foundation Policy for Robotic Manipulation}, 
      author={Rujia Yang and Geng Chen and Chuan Wen and Yang Gao},
      year={2025},
      eprint={2503.08950},
      archivePrefix={arXiv},
      primaryClass={cs.RO},
      url={https://arxiv.org/abs/2503.08950}, 
}

@misc{vjepa2,
      title={V-JEPA 2: Self-Supervised Video Models Enable Understanding, Prediction and Planning}, 
      author={Mido Assran and Adrien Bardes and David Fan and Quentin Garrido and Russell Howes and Mojtaba and Komeili and Matthew Muckley and Ammar Rizvi and Claire Roberts and Koustuv Sinha and Artem Zholus and Sergio Arnaud and Abha Gejji and Ada Martin and Francois Robert Hogan and Daniel Dugas and Piotr Bojanowski and Vasil Khalidov and Patrick Labatut and Francisco Massa and Marc Szafraniec and Kapil Krishnakumar and Yong Li and Xiaodong Ma and Sarath Chandar and Franziska Meier and Yann LeCun and Michael Rabbat and Nicolas Ballas},
      year={2025},
      eprint={2506.09985},
      archivePrefix={arXiv},
      primaryClass={cs.AI},
      url={https://arxiv.org/abs/2506.09985}, 
}

@misc{svp,
      title={Spatially Visual Perception for End-to-End Robotic Learning}, 
      author={Travis Davies and Jiahuan Yan and Xiang Chen and Yu Tian and Yueting Zhuang and Yiqi Huang and Luhui Hu},
      year={2024},
      eprint={2411.17458},
      archivePrefix={arXiv},
      primaryClass={cs.CV},
      url={https://arxiv.org/abs/2411.17458}, 
}

@misc{robograsp,
      title={RoboGrasp: A Universal Grasping Policy for Robust Robotic Control}, 
      author={Yiqi Huang and Travis Davies and Jiahuan Yan and Xiang Chen and Yu Tian and Luhui Hu},
      year={2025},
      eprint={2502.03072},
      archivePrefix={arXiv},
      primaryClass={cs.RO},
      url={https://arxiv.org/abs/2502.03072}, 
}

@misc{srgrasp,
      title={Spatial RoboGrasp: Generalized Robotic Grasping Control Policy}, 
      author={Yiqi Huang and Travis Davies and Jiahuan Yan and Jiankai Sun and Xiang Chen and Luhui Hu},
      year={2025},
      eprint={2505.20814},
      archivePrefix={arXiv},
      primaryClass={cs.RO},
      url={https://arxiv.org/abs/2505.20814}, 
}

\end{document}